\documentclass{article} 

\usepackage{iclr2021_conference,times}
\usepackage{subfiles}

\usepackage{amsmath,amsfonts,bm}

\def\eqref#1{equation~\ref{#1}}

\def\1{\bm{1}}

\DeclareMathAlphabet{\mathsfit}{\encodingdefault}{\sfdefault}{m}{sl}
\SetMathAlphabet{\mathsfit}{bold}{\encodingdefault}{\sfdefault}{bx}{n}

\usepackage{hyperref}
\usepackage{url}

\usepackage[ruled]{algorithm2e}
\usepackage{wrapfig}

\usepackage{graphicx}
\usepackage{subcaption}
\usepackage{float}

\usepackage[toc,page]{appendix}
\usepackage{makecell}

\usepackage{multirow}
\usepackage{color, colortbl}
\definecolor{LightGrey}{rgb}{0.90,0.90,0.90}
\definecolor{Grey}{rgb}{0.73,0.73,0.73}

\title{Zero-shot generalization using cascaded system-representations}

\author{\hfill Ashish Malik \\
\hfill Department of Mechanical Engineering\\
\hfill Punjab Engineering College\\
\hfill Chandigarh, 160012, India \\
\hfill ashishmalik.bemech14@pec.edu.in
}

\iclrfinalcopy 
\begin{document}

\maketitle

\begin{abstract}
	
	Behavior decompositions into primitives have been widely studied in hierarchical reinforcement learning and are known to improve generalization to changes in the environment. Conversely, does morphological decompositions into modular representations of agents lead to better generalization to changes in agent's morphology in the same environment? In this regard, we propose a novel approach based on recurrent neural networks which learns modular representations of robots in tandem with a higher level meta-policy. We demonstrate the effectiveness of our approach by learning common control policies for a variety of analogous robots with varied kinematics and dynamics. Our emphirical results show that the policies learned using our approach can achieve performance equivalent to expert policies (policies which are trained independently for each separate robot) on previously unseen analogous robots - even with different state and action dimensionalities.

\end{abstract}

\section{Introduction}

\begin{wrapfigure}[21]{r}{0.5\textwidth}
	\vspace{-0.5em}
	\begin{subfigure}{0.5\textwidth}
		\centering
		\includegraphics[width=1.0\linewidth]{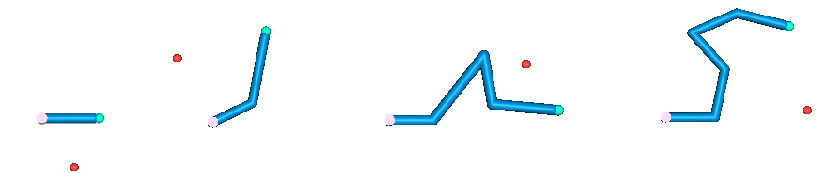}
		\caption{2-D Reachers}
	\end{subfigure}

	\begin{subfigure}{0.5\textwidth}
		\centering
		\includegraphics[width=1.0\linewidth]{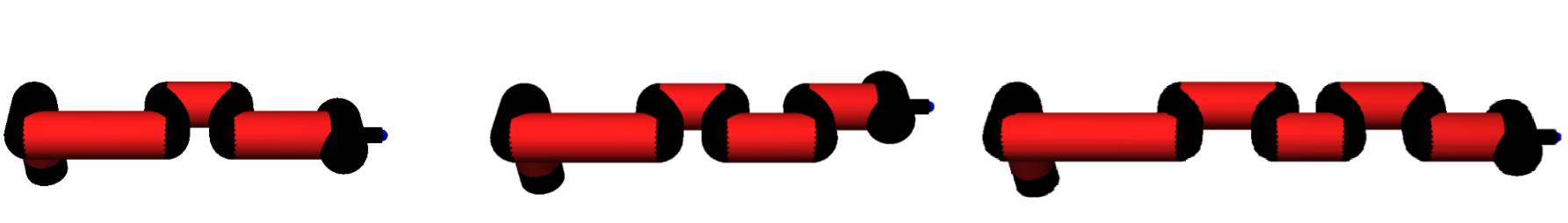}
		\caption{Manipulators}
	\end{subfigure}

	\begin{subfigure}{0.5\textwidth}
		\centering
		\includegraphics[width=1.0\linewidth]{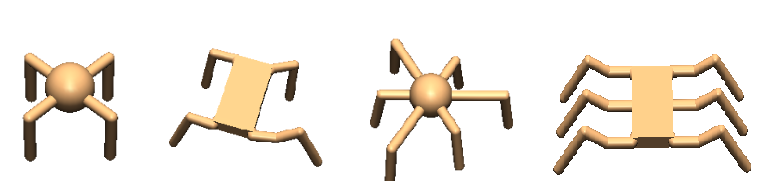}
		\caption{Crawling robots}
	\end{subfigure}

	\caption{Examples of various domains of control problems for which we aim to develop a universal policy.}
	\label{environmentTypes}
\end{wrapfigure}

Deep reinforcement learning (DRL), has recently gained unprecedented success in sensorimotor control, finance, health-care, etc. However, widespread real-world adaptation of deep reinforcement learning still presents a few challenges. One of these challenges is lack of generalizability/transferability of learned policies. The performance of learned policies diminishes when properties of the agent, environment or tasks changes to regimes not encountered during training. Real world designs often do not follow a standard which raises the need to learn policies that are invariant to design choices and perform well across different operating environments. In robotics this problem manifests itself in form of morphologically different robots with vastly dissimilar hardware specifications and operational characteristics being used for similar operations (Eg. Rethink robotic’s Sawyer, Universal robotic’s UR16, etc). Currently, this necessitates the need for developing and maintaining separate hardware and morphology dependent controllers. Efforts have been made to mitigate this dependence using methods such as dynamics randomization \citep{peng2018sim} forming kernal mismatch models \cite{rai2019using} and learning variational policy embeddings \cite{atnekvist2019vpe} etc.. Nonetheless, learning invariant policies for different robots still remains challenging.

\begin{wrapfigure}[17]{l}{0.5\textwidth}
	\begin{subfigure}{0.52\textwidth}
		\centering
		\includegraphics[width=1.0\linewidth]{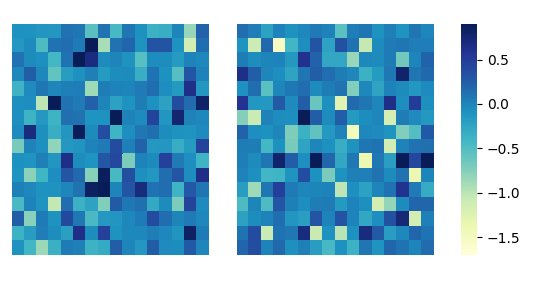}
	\end{subfigure}
	\caption{Abstract representations formed by the penultimate layer of networks used for training inverted pendulum agent with different seeds. Both used the same learning algorithm, hyperparameters and achieved equivalent performance.}
	\label{representationsHeatmap}
\end{wrapfigure}

We begin with a hypothesis that the abstract-representations learned by a neural network using conventional DRL are one of the many possible representations which could lead to similar policy performance (see fig. \ref{representationsHeatmap}). Each of these representations entangle explanatory features required for learning optimal policy differently. The exact representations learned depend upon the network's architecture, its initialization \citep{frankle2018lottery} and the learning process. Among these representations, some better capture the morphological and hardware features of robots and are thus better suited for generalizablity. Our proposed approach enforces learning of such representations. They are learned in tandem with the meta policy to ensure they are effective for the task. Better representations improves policy's generalizability which in-turn improves quality of representations for learning policy. 

Our work advances the idea proposed by \citet{chen2018hardware} to learn invariant policies by conditioning them on varying morphologies and hardwares. But unlike their approach which uses simple feed-forward networks, we propose a new recurrent neural architecture which we name \textbf{Cas}caded model \textbf{net}work (CASNET). Our CASNET architecture is uses recurrent neural networks (RNNs) to better capture the structural features of different robots in their learned representations. The architecture is composed of 3 identifiable modules. First consists of a sequence of recurrent cells to generate morphologically dependent vector-representations. Second is a ordinary feed-forward network to learn the meta-policy and third consist of another sequence of recurrent cells to decode output of the meta-policy into actions corresponding to the robot's morphology. CASNET is compatible with any reinforcement learning algorithm that uses temporal difference updates for state or action value functions.CASNET also provide other important advantages over \citet{chen2018hardware} method which we discuss ahead in the paper. Our experimental results show that universal policies which generalize over a variety of robot morphologies and hardwares can be learned using CASNET. These policies also achieve zero-shot transfer to previously unseen robots.

\section{Formalization}

	Let $\mathbb{S} := \{(\mathbf{s_t^a}, \mathbf{s_t^e})\}$ be the joint state of the agent and its operating environment at any arbitrary time-step $t$. Agent's observation, actions and rewards at time $t$ are $\mathbf{o}_t \in \mathbb{O}$, $\mathbf{a}_t \in \mathbb{A}$ and $r_t \in \mathbb{R}$, respectively. We define a degree of variation $v$ as an independent variable which is represented by an arbitrary element of $\mathbf{s_t^a} \in \mathbb{S}$, i.e.,  $v_i := \mathbf{s_t^a}[i]$. Finally, we assume access to a set of $N$ different training environments\footnote{Note that we are using the term \textit{environment} both for the operating environments that the agent interacts with and also for simulation environment which consist of the agent and its operating environment} $\boldsymbol{\varepsilon}_{\text{tr}} := \{e^\mathbb{D}_i\}^N_{i=1}$ drawn from a predetermined domain $\mathbb{D}$ during training. The domain $\mathbb{D}$ is determined as: 
	\begin{equation}
		\mathbb{O}^{e \sim \mathbb{D}} \subseteq \mathbb{S} \quad \big| \quad (\mathbf{s_t^a})_i \in [\min(v_i), \max(v_i)]
	\end{equation}

	Therefore, $\mathbb{D}$ is a set of all possible environments for which values of $v$ lies between specified bounds. Let $\{(\mathbf{o_t}^e, \mathbf{a_t}^e, r_t^e)_{t=0}^{N}\}_{e\sim \boldsymbol{\varepsilon}_{\text{tr}}}$ be the data collected during training. Our goal is to use this data to formulate a predictor $P : \mathbb{O} \rightarrow \mathbb{A}$ that performs well across all $e \sim \mathbb{D}$. Emulating \citet{ahuja2020invariant}\footnote{Their work deals with risk minimization across environments which can be trivally modified to returns maximization across different environments.}, we achieve this objective by learning a data-representation $\Psi : \mathbb{O} \rightarrow \Upsilon \subseteq \mathbb{R}$ that elicits an invariant predictor $w \circ \Psi : \Upsilon \rightarrow \mathbb{A}$, which simultaneously maximize returns across all environments. This can be phrased as the following constraint optimization problem:
	\begin{equation}
	\label{optimizationOriginal}
	\begin{aligned}
	& \underset{\Psi \in \mathbb{H}_\Psi, w \in \mathbb{H}_w}{\max} \sum_{e \sim \boldsymbol{\varepsilon}_{\text{tr}}} \mathbb{E} \big[ G^e(w \circ\Psi) \big] \\
	\text{subject to: } & w \in \text{arg } \underset{w' \in \mathbb{H}_w}{\max} \mathbb{E} \big[ G^e(w' \circ \Psi) \big],\forall e \in \boldsymbol{\varepsilon}_{\text{tr}}
	\end{aligned}
	\end{equation}
	Here, $\mathbb{H}_\Psi$ and $\mathbb{H}_w$ are the set of all mappings $\Psi$ and predictors $w$ respectively. $G^e = \sum_{t=0}^{n} \gamma^{t} r_t$ are the returns (discounted or otherwise over finite or infinite horizon) recieved from the environment $e$. As pointed by \citet{ahuja2020invariant}, the optimization problem \ref{optimizationOriginal} is difficult to solve directly and an alternate game theory based characterization can be used instead. Assume each environment has its own predictor $w^e \in \mathbb{H}_w$ and their ensemble constructs an overall predictor $\bar{w}:\Upsilon \rightarrow \mathbb{A} := \frac{1}{|\boldsymbol{\varepsilon}_{\text{tr}}|}\sum_{j=1}^{|\boldsymbol{\varepsilon}_{\text{tr}}|}w^j(\psi)$, where $\psi \in \Psi$ . The new optimization problem then is:
	\begin{equation}
	\label{optimizationGame}
	\begin{aligned}
	& \underset{\Psi \in \mathbb{H}_\Psi, \bar{w} \in \mathbb{H}_w}{\max} \sum_{e \sim \boldsymbol{\varepsilon}_{\text{tr}}} \mathbb{E} \big[ G^e(\bar{w} \circ\Psi) \big] \\
	\text{subject to: } & w^e \in \text{arg } \underset{w'^e \in \mathbb{H}_w}{\max} \mathbb{E} \Big[ G^e\Big((w'^{e} + \sum_{j\neq e}w^j) \circ \Psi\Big) \Big],\forall e \in \boldsymbol{\varepsilon}_{\text{tr}}
	\end{aligned}
	\end{equation}
	The above constraints can be equivalently stated as a pure Nash equlibria of a game as:
	\begin{equation}
	\label{optimizationNash}
	\begin{aligned}
	& \underset{\Psi \in \mathbb{H}_\Psi, \bar{w}}{\max} \sum_{e \sim \boldsymbol{\varepsilon}_{\text{tr}}} \mathbb{E} \big[ G^e(\bar{w} \circ\Psi) \big] \\
	\text{subject to: } & \mathbb{E} \Big[ G^e\Big( \frac{1}{|\boldsymbol{\varepsilon}_{\text{tr}}|} \big[w^e + \sum_{j \neq e} w^j\big] \circ \Psi \Big) \Big] \\
	&\geq \mathbb{E} \Big[ G^e\Big( \frac{1}{|\boldsymbol{\varepsilon}_{\text{tr}}|} \big[w'^e + \sum_{j \neq e} w^j\big] \circ \Psi \Big) \Big], \forall w'^e \in \mathbb{H}_w, \forall e \in \boldsymbol{\varepsilon}_{\text{tr}}
	\end{aligned}
	\end{equation}
	This is a non-zero sum continuous game that is played between $|\boldsymbol{\varepsilon}_{\text{tr}}|$ players. Each player corresponds to $e \in \boldsymbol{\varepsilon}_{\text{tr}}$ and their set of actions are choosing $w^e \in \mathbb{H}_w$. Each player is trying to maximize its own utility function $\mathbb{E}\big(R^e(\bar{w} \circ \Psi)\big)$, given a representation $\Psi$. Finding Nash equilibrium for continuous games is a challenging task for which several heuristic schemes have been proposed. Yet none of them are guaranteed to compute pure or mixed equilibrium for non-zero sum games apart from some special cases. Moreover, learning a separate $w^e$ for each environment is in-efficient, so, we instead learn $\bar{w}$ directly. Let $\bar{w}$ be modeled with parameters $\boldsymbol{\theta}_{\bar{w}}$. During training, each player competes to move $\bar{w}$ towards $w^e$ (this can be understood as a modified \textit{Best response dynamics} strategy \citet{barron2010best} for finding Nash equilibrium). This characterization changes \ref{optimizationNash} to:
	\begin{equation}
	\label{optimizationCASNET}
	\begin{aligned}
	& \underset{\Psi \in \mathbb{H}_\Psi, \bar{w}}{\max} \sum_{e \sim \boldsymbol{\varepsilon}_{\text{tr}}} \mathbb{E} \big[ G^e(\bar{w} \circ\Psi) \big] \\
	\text{subject to : }
	&\mathbb{E} \Big[ G^e \big( \bar{w} \circ \Psi \big) \Big] 
	\geq \mathbb{E} \Big[ G^e\big( \bar{w'} \circ \Psi \big) \Big], \forall \bar{w'} \in \mathbb{H}_w, \forall e \in \boldsymbol{\varepsilon}_{\text{tr}}
	\end{aligned}
	\end{equation}
	Characterization \ref{optimizationCASNET} is mathematically and computationally more tractable compared to \ref{optimizationNash}. It additionally allows us to train $\boldsymbol\theta_{\bar{w}}$ using heterogeneous data batches which consist of data samples generated by different environments. To satisfy the constraint in \ref{optimizationCASNET}, the learned representation $\Psi$ should not include any features which may lead to spurious correlations to successful or unsuccessful actions for any subsets of environments. Thus, the representation learner will learn to ignore distractors in the input data when trained in tandem with the invariant predictor. Intuitively, the resulting invariant policy finds causes of successful actions across environments and thus improves generalizability.

\section{Cascaded model network}
	
	\begin{figure}[ht]
		\centering
		\includegraphics[width=1.0\linewidth]{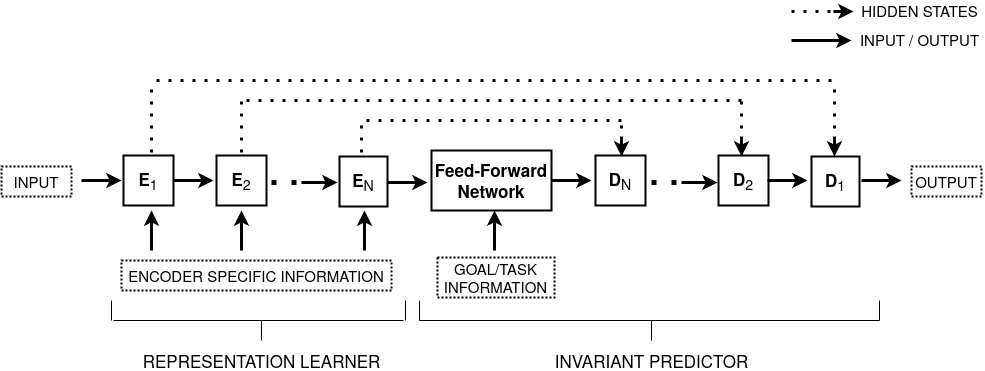}
		\caption{The CASNET architecture. Here E and D stands for recurrent neural network based encoder and decoder modules respectively. The cascade of lower to higher level representations leads to learning causality.}
		\label{casnetArch}
	\end{figure}
	The proposed neural architecture is shown in fig. \ref{casnetArch}. It is composed of a sequence (cascade) of encoders which are recurrent cells, a feed-forward network that acts as meta-policy and finally another sequence of recurrent cells acting as decoders. The idea behind CASNET architecture is the hypothesis that the world is inherently compositional. It can be expressed as compositions of small sets of primitive mechanisms and structures \citet{parascandolo2018learning}. Complexity in the real world emerges through combinations of these primitives. By extension, analogous systems are composed of analogous subsystems assembled in different configurations. These sub-systems and their specific configurations determines a system's distinctiveness and thus its behavior. As it is evident that humans can combine simple skills to perform highly complicated tasks, it might be a useful inductive bias for the learning models to generate complex representations of higher order systems by combining representations of lower order simpler systems. A representative example of complex higher order systems being made up of simpler systems is rotary wing UAVs. They are made up of made up of various number of propellers attached at different locations w.r.t. the center of mass on the body. These propellers can have similar or different aerodynamic properties. Another example can be crawling robots which are made up of various number of legs (first order sub-system). These legs are themselves made up of different number of actuator link pairs (second order sub-system).

	Input to the CASNET architecture are agent's observations decomposed into sets of lowest order subsystem's observations. The first RNN cell uses this input to form sets of representations of higher-order sub-systems which are then passed as input to the next RNN cell. This sequence is followed until representation of the entire agent is achieved (Representation learner in fig. \ref{casnetArch}). This representation is then augmented with any external observation such as goal or environment's observation and passed to meta-policy's network. Output of the meta-policy is decoded in inverse order as that of encoders using another sequence of RNNs cells to finally output action values. Input to a decoder is output of the previous decoder (or meta-policy) combined with the hidden-states of the corresponding encoder. These hidden states enhance the input to the decoder by infusing additional sub-system specific morphological information. Such decomposition along with the use of RNN cells makes CASNET's design modular that can work with arbitrary number of subsystems and action dimensions. The CASNET architecture forces the learner to discover high-level representations that better capture the composition (structural or causal) of the agents and thus leads to better generalization results. Because of this modular design, CASNET does not assume identical dimensionality of action or observation spaces for environments sampled from the same domain. On the other hand, the dimensionality of the most primitive subsystem's observations and actions are assumed constant across the entire domain. However, this assumption is much less restrictive than the previous one.

	\textbf{Objective function design}. The objective function design for each environment should be normalized to be independent of the number of subsystems of any order. This normalization prevents overfitting by ensuring that the learning process is not dominated by some sub-set of training environments which have higher or lower number of subsystems than the average. Let $L^e$ be the normalized objective function for any $e \sim \mathbb{D}$. Then objective for training CASNET is:
	\begin{equation}
	L_{\text{casnet}} \propto \sum_{e \sim \boldsymbol{\epsilon}_\text{tr}} L^e
	\end{equation}

	\textbf{Sequence of observations and learned representations}. Subsystems can have serial (actuator-link pairs in manipulators) or parallel (legs in crawling robots) configurations. Output of the RNN cells is sensitive to the sequence of the input data. Therefore, a sequencing rule should be established for parallel configurations and should be followed for the entire domain. Representations / observations of these parallel sub-systems should be augmented with configuration information (encoder specific information in fig. \ref{casnetArch}) so that the policy is aware of the exact morphology/design of the agent.

\section{Experiments}
	
	We used 3 separate domains of robotics to investigate the generalization ability of our CASNET architecture: Planer reachers, crawling robots, and 3-D manipulators. Our experiments were conducted using OpenAI gym \citep{brockman2016openai} and Mujoco physics engine \citep{todorov2012mujoco}. Overview of these domains are given ahead are discussed in details in the Appendix. DOVs and their limits for each domain is mentioned in table \ref{envDetails}.

	\begin{wraptable}[16]{r}{0.44\textwidth}
		\fontsize{8}{7.2}\selectfont
		\vspace{-1.4em}
		\caption{DOVs for domains used in our experiments. Values are in SI units wherever applicable.}
		\label{envDetails}
		\centering
		\begin{tabular}{|c|c|} 
			\hline
			\rowcolor{Grey}
			DOV & Limits\\
			\hline\hline
			\rowcolor{LightGrey}
			\multicolumn{2}{|c|}{Reacher domain}\\
			\hline
			DOF & [1,5]\\
			\hline
			Link-lengths & [0.07,0.17]\\ 
			\hline
			\rowcolor{LightGrey}
			\multicolumn{2}{|c|}{Crawler domain}\\
			\hline
			DOF per leg & 2 or 3\\
			\hline
			Num. legs & 4 or 6\\
			\hline
			Legs symmetry & Bilateral/Radial/Irregular\\
			\hline
			Joint angle limits & [0.7-1.6]\\
			\hline
			Link-lengths & [0.07,0.17]\\
			\hline
			\rowcolor{LightGrey}
			\multicolumn{2}{|c|}{Manipulator domain}\\
			\hline
			DOF & [5, 6, 7]\\
			\hline
			Link-lengths & 3 $\pm$ 25$\%$\\ 
			\hline
			Joint damping & 30 $\pm$ 25$\%$\\ 
			\hline
			Joint friction & 10 $\pm$ 25$\%$\\ 
			\hline
			Actuator torque & 80 $\pm$ 25$\%$\\ 
			\hline
		\end{tabular}
	\end{wraptable}

	\textbf{Reachers}: 15 unique simulation environments were created with the goal of reaching a randomly located (but reachable) goal. There were 3 environments per DOF of which 1 was used for training and other 2 for testing.

	\textbf{Crawling robots}: 56 unique simulation environments were created with goal to walk in the forward direction (+y axis). 46 of these 56 environments have robot designs with variations along a single DOV (apart from total number of legs) while 10 have variations along multiple DOVs. 37 environments (including all 10 with variations along multiple DOVs) were used for testing and the rest for training.

	\textbf{Manipulators}: 270 environments were created by randomly sampling DOV values between specified limits (90 environments per DOF). 10 environments per DOF were used for testing. Each environment followed a sparse reward setting of +1 reward for reaching the goal and 0 otherwise.

	We compare our proposed CASNET architecture to the following baselines:

	\begin{itemize}
		\item Expert policies: Expert policies refer to policies learned for a single environment using an appropriately sized feed forward network.
		\item Implicitly encoded hardware conditioned policies (HCP-I): \citet{chen2018hardware} showed that HCP-I can achieve state of the art performance in generalizing over a variety of robots.
	\end{itemize}

	\subsection{Generalization experiments}
		\begin{wrapfigure}[15]{r}{0.6\textwidth}
		\vspace{-1em}
		\begin{subfigure}{0.29\textwidth}
			\centering
			\includegraphics[width=1.0\linewidth]{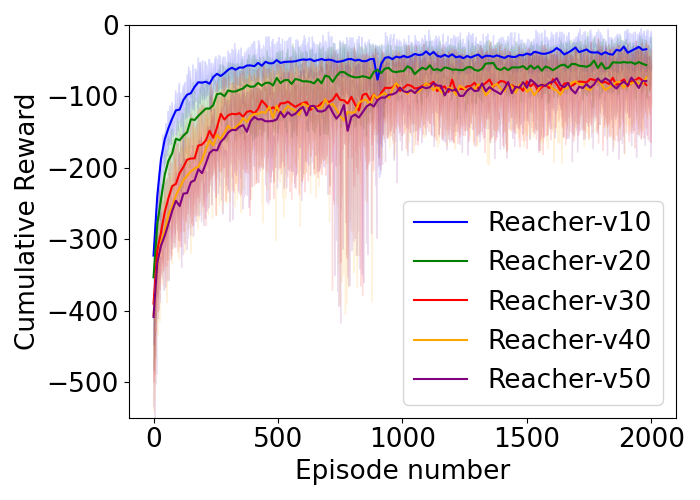}
			\caption{CASNET Policy}
			\label{casnetLearning}
		\end{subfigure}
		\begin{subfigure}{0.29\textwidth}
			\centering
			\includegraphics[width=1.0\linewidth]{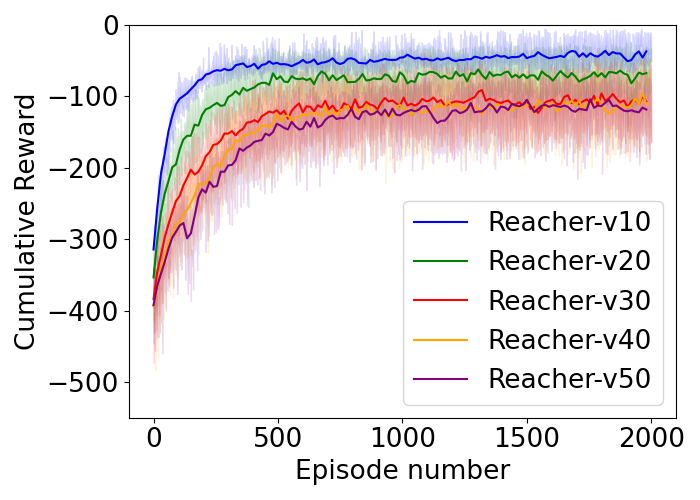}
			\caption{Expert Policies}
			\label{expertLearning}
		\end{subfigure}
		\caption{CASNET and expert policies learned for planer manipulators. Note that the expert policies are trained independently but are shown together in fig. \ref{expertLearning} for facilitating comparison with the CASNET policy shown in fig. \ref{casnetLearning}.}
		\label{reacherLearning}
		\end{wrapfigure}

		In this section we compare the generalization performance of out CASNET approach with baselines. Every experiment is conducted with 3 random seed value (1,2 and 3) until convergence. Hyperparameter values used are given in Appendix.

		\textbf{Reacher domain}: Proximal policy optimization (PPO) algorithm \citep{schulman2017proximal} with generalized advantage estimates (GAE) \citep{schulman2015high} was used to train policies for Reacher domain. The CASNET agent used a single pair of encoder-decoder pair. The learning performance of CASNET policy is compared to expert policies in fig \ref{reacherLearning}. Table \ref{reachersPerformance} compares the final performance of CASNET policy with baselines on test environments.

		\textbf{Crawler robots domain}: Soft actor critic (SAC) algorithm \citep{haarnoja2018soft} was used to train policies for Crawling robots. The CASNET agent used 2 encoder-decoder pairs, one for encodings and decoding leg states and other for the whole robots. Performance comparison of CASNET and baselines on test environments is summarized in table \ref{crawlersPerformance}.

		\textbf{Manipulator domain}: Deep deterministic policy gradient (DDPG) \citep{silver2014deterministic} with Hindsight experience replay (HER) \citep{andrychowicz2017hindsight} for training policies for manipulator domain. The CASNET architecture used is similar to the one used for Reacher domain. Table \ref{manipulatorsPerformance} compares the performance of the CASNET policy against the baselines. Nomenclature Manipulator\_Yx means testing environments with Y DOFs.

		\begin{wraptable}[8]{l}{0.45\textwidth}
		\fontsize{7}{7.2}\selectfont
		\vspace{-1.4em}
		\caption{Performance comparison: Manipulator domain}
		\label{manipulatorsPerformance}
		\centering
		\setlength\tabcolsep{1.5pt} 
				\begin{tabular}{|c|ccc|ccc|ccc|} 
				\hline
				\multirow{2}{3em}{\hspace{0.4em}Name} & \multicolumn{9}{c|}{Success-ratio $\pm$  Standard dev.}\\
				\cline{2-10}
				&\multicolumn{3}{c|}{Expert} & \multicolumn{3}{c|}{CASNET} & \multicolumn{3}{c|}{HCP-I}\\
				\hline\hline
				Manipulator-v5x & \textbf{0.91} & $\pm$ & \textbf{0.14} & 0.84 & $\pm$ & 0.22 & 0.76 & $\pm$ & 0.20\\
				\hline
				Manipulator-v6x & 0.79 & $\pm$ & 0.17 & \textbf{0.83} & $\pm$ & \textbf{0.18} & 0.73 & $\pm$ & 0.20\\
				\hline
				Manipulator-v7x & \textbf{0.76} & $\pm$ & \textbf{0.18} & 0.67 & $\pm$ & 0.34 & 0.59 & $\pm$ & 0.25\\
				\hline\hline
				Cumulative & \textbf{0.82} & $\pm$ & \textbf{0.18} & 0.78 & $\pm$ & 0.27 & 0.70 & $\pm$ & 0.23\\
				\hline
				\end{tabular}
		\end{wraptable}

		For our selected domains, we see that CASNET regularly performs considerably better than HCP-I. It even performs better than expert policies in significant number of cases. This signifies that CASNET architecture can learn the underlying causal structures and morphological features which help in learning robust policies. The learned policies consistently perform well across the selected domain.

		\begin{table}[ht]
		\fontsize{7}{7.2}\selectfont
		\parbox{.44\linewidth}{ 
		\centering
		\caption{Performance comparison: Reacher domain}
		\label{reachersPerformance}
		\setlength\tabcolsep{1.5pt}
		\begin{tabular}{|c|ccc|ccc|ccc|}
				\hline
				\multirow{2}{3em}{\hspace{0.4em}Name} & \multicolumn{9}{c|}{Avg. rewards $\pm$  Standard dev.}\\
				\cline{2-10}
				&\multicolumn{3}{c|}{Expert} & \multicolumn{3}{c|}{CASNET} & \multicolumn{3}{c|}{HCP-I}\\
				\hline\hline
				Reacher-v11 & -68.1 & $\pm$ & 3.2 & \textbf{-48.6} & $\pm$ & \textbf{5.5} & -62.1 & $\pm$ & 5.5\\
				\hline
				Reacher-v12 & -42.7 & $\pm$ & 4.0 & \textbf{-33.5} & $\pm$ & \textbf{2.3} & -42.0 & $\pm$ & 1.9\\
				\hline
				Reacher-v21 & -72.4 & $\pm$ & 1.5 & \textbf{-58.3} & $\pm$ & \textbf{2.7} & -65.1 & $\pm$ & 3.6\\
				\hline
				Reacher-v22 & -80.8 & $\pm$ & 2.2 & \textbf{-65.7} & $\pm$ & \textbf{3.3} & -72.7 & $\pm$ & 4.6\\
				\hline
				Reacher-v31 & -85.0 & $\pm$ & 3.1 & \textbf{-65.1} & $\pm$ & \textbf{1.7} & -77.7 & $\pm$ & 5.4\\
				\hline
				Reacher-v32 & -99.1 & $\pm$ & 2.4 & \textbf{-78.6} & $\pm$ & \textbf{4.7} & -91.4 & $\pm$ & 5.2\\
				\hline
				Reacher-v41 & -102.3 & $\pm$ & 3.0 & \textbf{-74.6} & $\pm$ & \textbf{2.3} & -90.5 & $\pm$ & 4.1\\
				\hline
				Reacher-v42 & -106.0 & $\pm$ & 3.9 & \textbf{-75.4} & $\pm$ & \textbf{4.0} & -102.2 & $\pm$ & 8.7\\
				\hline
				Reacher-v51 & -135.8 & $\pm$ & 4.3 & \textbf{-95.4} & $\pm$ & \textbf{0.6} & -116.5 & $\pm$ & 2.6\\
				\hline
				Reacher-v52 & -130.2 & $\pm$ & 1.3 & \textbf{-95.1} & $\pm$ & \textbf{2.3} & -116.3 & $\pm$ & 5.3\\
				\hline\hline
				Cumulative & -92.2 & $\pm$ & 27.2 & \textbf{-69.0} & $\pm$ & \textbf{18.6} & -83.6 & $\pm$ & 23.5\\
				\hline
				\end{tabular}
		}
		\hfill
		\parbox{.54\linewidth}{
		\centering
		\caption{Performance comparison: Crawler-robots domain}
		\label{crawlersPerformance}
		\setlength\tabcolsep{1.5pt}
		\begin{tabular}{|c|ccc|ccc|ccc|} 
				\hline
				\multirow{2}{5em}{\hspace{0.4em}Robot-type} & \multicolumn{9}{c|}{Avg. rewards $\pm$ Standard dev.}\\
				\cline{2-10}
				&\multicolumn{3}{c|}{Expert} & \multicolumn{3}{c|}{CASNET} & \multicolumn{3}{c|}{HCP-I}\\
				\hline\hline
				Hexapod-v1x		&		961.2	&$\pm$	&	66.1		&	\textbf{1003.9}	&$\pm$	&	\textbf{6.5}		&	977.4	&$\pm$	&	46.9\\
				\hline
				Hexapod-v2x		&		648.7	&$\pm$	&	435.3		&	\textbf{694.8}	&$\pm$	&	\textbf{298.8}	&	543.3	&$\pm$	&	428.7\\
				\hline
				Hexapod-v3x		&		976.6	&$\pm$	&	34.3		&	\textbf{1006.9}	&$\pm$	&	\textbf{6.5}		&	1006.4	&$\pm$	&	5.6\\
				\hline
				Hexapod-v4x		&		132.3	&$\pm$	&	199.1		&	\textbf{992.3}	&$\pm$	&	\textbf{19.2}	&	458.9	&$\pm$	&	483.5\\
				\hline
				Hexapod-v5x		&		\textbf{877.4}	&$\pm$	&	\textbf{178.1}		&	804.9	&$\pm$	&	256.7	&	650.9	&$\pm$	&	437.3\\
				\hline
				Quadruped-v1x	&		983.0	&$\pm$	&	110.2		&	997.4	&$\pm$	&	4.4		&	\textbf{1002.8}	&$\pm$	&	\textbf{7.7}\\
				\hline
				Quadruped-v2x	&		417.8	&$\pm$	&	495.5		&	\textbf{660.3}	&$\pm$	&	\textbf{247.1}	&	599.4	&$\pm$	&	447.9\\
				\hline
				Quadruped-v3x	&		287.6	&$\pm$	&	371.2		&	\textbf{723.2}	&$\pm$	&	\textbf{420.7}	&	693.4	&$\pm$	&	418.6.1\\
				\hline
				Quadruped-v4x	&		990.2	&$\pm$	&	34.9		&	1000.4	&$\pm$	&	3.7		&	\textbf{1007.0}	&$\pm$	&	\textbf{16.5}\\
				\hline
				Quadruped-v5x	&		807.7	&$\pm$	&	281.9		&	843.2	&$\pm$	&	203.0	&	\textbf{879.6}	&$\pm$	&	\textbf{293.4}\\
				\hline\hline
				Cumulative		&		720.1	&$\pm$	&	399.0		&	\textbf{860.5}	&$\pm$	&	\textbf{250.8}	&	774.0	&$\pm$	&	388.8\\
				\hline
				\end{tabular}
		}
		\end{table}

	\subsection{Learned data-representations}
\begin{figure}[ht]
			\centering
			\begin{subfigure}{0.49\textwidth}
				\centering
				\includegraphics[width=0.8\linewidth]{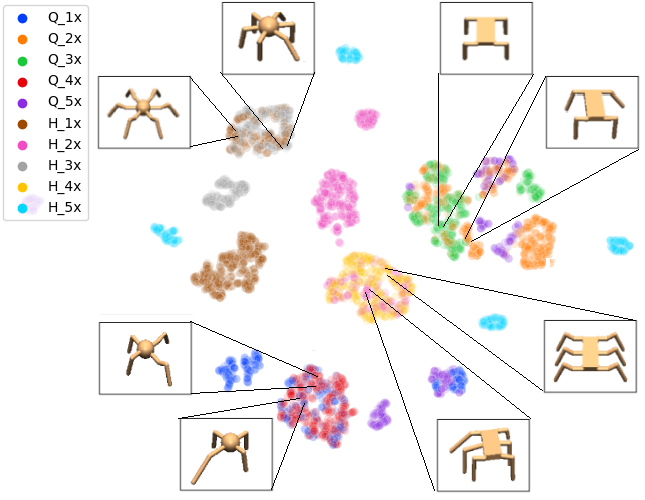}
				\caption{Learned crawler robots representations}
				\label{crawlerRepresentations}
			\end{subfigure}
			\begin{subfigure}{0.49\textwidth}
				\centering
				\includegraphics[width=0.8\linewidth]{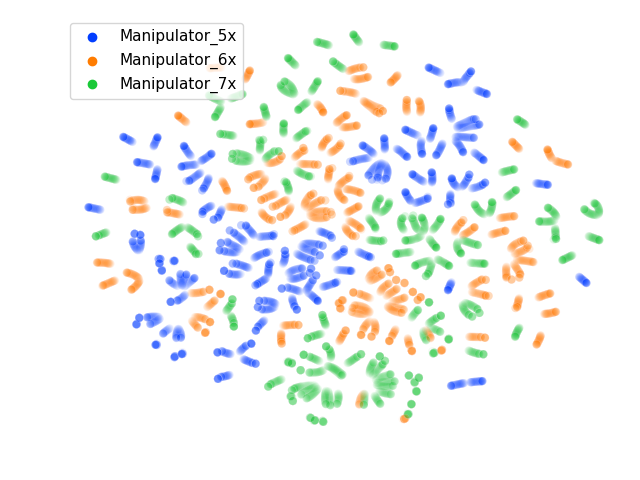}
				\caption{Learned manipulator representations}
				\label{manipulatorRepresentations}
			\end{subfigure}
			\caption{t-SNE \citep{maaten2008visualizing} visualizations of learned representations}
			\label{learnedRepresentations}
		\end{figure}

		The performance of the learned invariant policy is dependent on the quality of learned representations and how well the capture and differentiate features of different morphologies. The verify the characteristics of representations learned by CASNET's policy, we plot them (via-dimensionality reduction) in figure \ref{learnedRepresentations}.

		We observe that representations form distinct clusters based upon DOVs. Representations of crawling robots with multiple DOVs are spread among representations of robots with single DOVs. Also representations of similar morphologies (ex. H\_1x and H\_3x) often overlap. This indicates that CASNET can also captures similarities between different morphologies which is essential for better generalization.

	\subsection{Out of distribution generalization and retraining}
		\begin{figure}[ht]
			\includegraphics[width=1.0\linewidth]{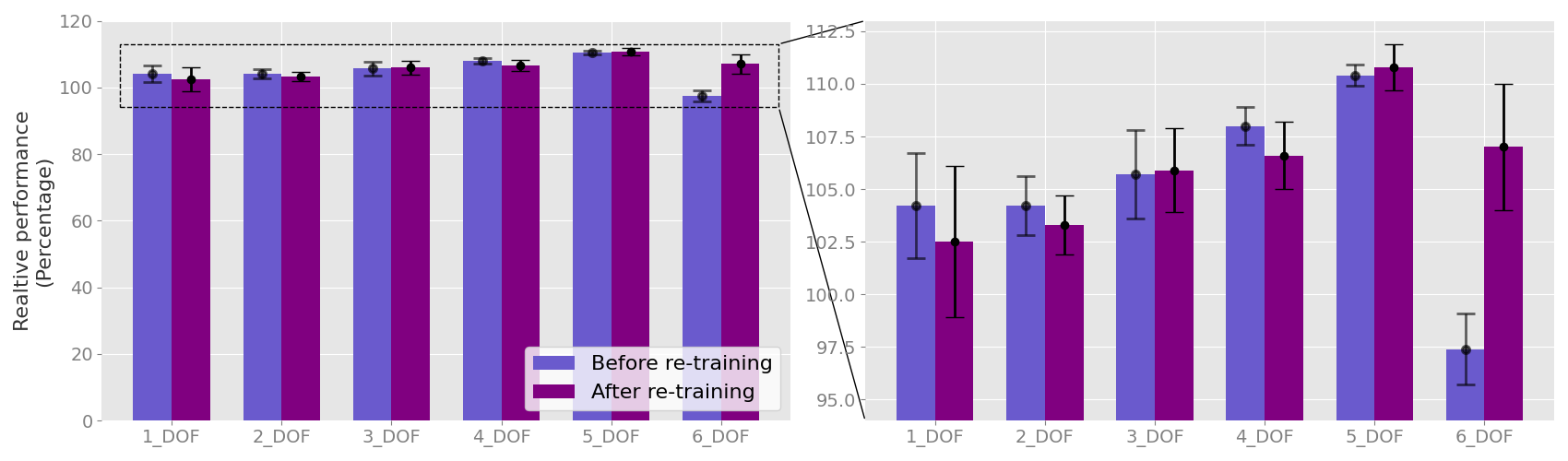}
			\caption{CASNET's performance before and after re-training for 6-DOF reachers. Here 100 percent corresponds to average performance of expert policies while 0 percent signifies performance of random actions.}
			\label{retrainingCASNET}
		\end{figure}

		Modular structures that capture the dynamics across environments have been shown to lead to better generalization and robustness to changes \citet{goyal2019recurrent}. In this sub-section,we determine how well the CASNET agents can capture the underlying causal mechanism across environments drawn from distribution outside the training domain. Additionally, as RNNs can take inputs of arbitrary size, a trained CASNET agent should be able to scale to new input and output sizes. For robotics, these mean that a learned CASNET policy, in theory, should be extendable to new environments with relatively relaxed DOV limits than the ones used to train the agent (including environments with different actions or state dimensionalities) with no or minimal retraining. Re-training, if required, can even be performed with only new morphologies thereby alleviating the need to store the training data and/or morphologies.

		\begin{wrapfigure}[16]{r}{0.67\textwidth}
			\vspace{-1.2em}
			\begin{subfigure}{0.67\textwidth}
				\centering
				\includegraphics[width=1.0\linewidth]{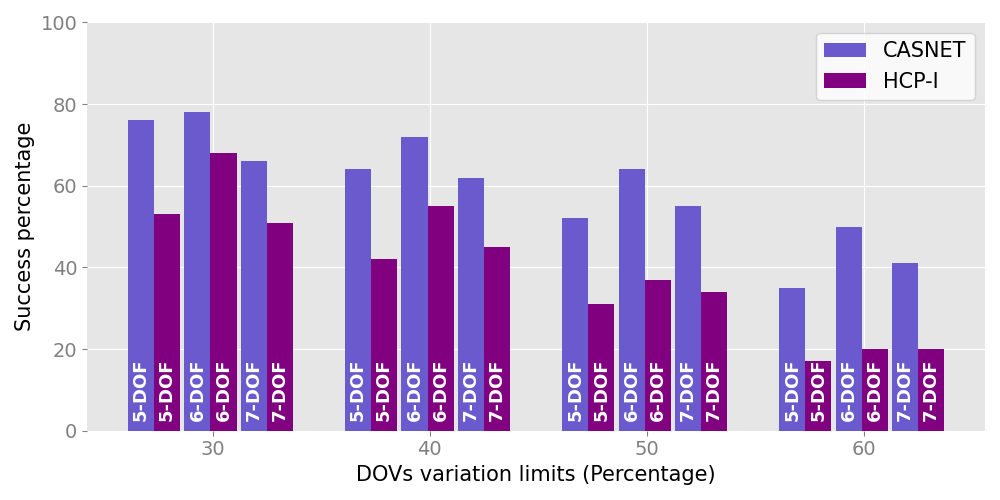}
				\end{subfigure}
			\caption{Performance on robots drawn from outside the training domain.}
			\label{outDistExp}
		\end{wrapfigure}

		3 new Reacher environments were created with 6-DOFs each while keeping other DOV limits the same to test how well the trained CASNET agent can perform with new state and action dimensionalities. The new environments had the same observation and action space composition as the previous environments (except for the addition of a new link). The performance of the CASNET policy on these new environment before and after retraining is shown in fig. \ref{retrainingCASNET}. The average retraining loss is 1.27\% across previous environments. Re-training the agent was done using a single 6-DOF Reacher and only used 1.5\% of the training data compared to training of the initial CASNET policy. The Hardware conditioned policies method cannot offer such flexibility due to its use of simple multi-layered perceptron and would thus require re-training from scratch.

		To test how well the CASNET agent perform on environments drawn from out of training distribution domain DOV limits were gradually increased from 25\% (limit used for training). The generalization results are shown in fig. \ref{outDistExp}. For each new DOV limit instance, 30 new environments were created and result are average of 100 episodes per DOF. The CASNET agents consistently performs considerably better than HCP-I agent with generalization performance difference increasing with increasing limits.

\section{Conclusion and future works}
	
	In summary, we introduced a new neural architecture named CASNET which can learn policies that generalize to analogous environments/systems. CASNET is compatible with any temporal difference based reinforcement learning method. We tested CASNET for learning control policies for 3 different and popular domains of robotics: Planer reachers, crawling robots and 3D Manipulators using state of the art on and off-policy learning algorithms. The on-par performance of the general policies learned using CASNET with expert policies trained individually for separate robot models demonstrate that the final performance of the CASNET policies are bound by the learning algorithm instead of the proposed framework. We also tested CASNET agent's ability to generalize to out-of-distribution data and established that the trained agent can capture the underlying causal mechanisms across different environments/robots.

\section{Related Works}

	Transfer learning has been a major focus of the reinforcement learning and robotics community. Extensive studies have been conducted in transferring policies from simulations to the real world \citep{peng2018sim, tobin2017domain, barrett2010transfer, zhang2016modular, zhang2017sim, van2019sim}, across tasks \citep{fu2016one, pinto2017learning} and dynamics \citep{rajeswaran2016epopt, mandlekar2017adversarially}. However, this current work is more closely related to transferring skills amongst multiple agents. Notable works in this direction include \citet{devin2017learning, hu2019skill, gupta2017learning, helwa2017multi, bocsi2013alignment, chen2018hardware}.

	\citet{bocsi2013alignment} used transfer learning to learn dynamics models for manipulators. They used the data generated from different experiments and robot models to obtain low dimensional manifolds.Subsequently, they used their proposed algorithm to find an isomorphism between these manifolds to transfer tasks across them. \citet{helwa2017multi} through the study of single-input single-output systems showed that these isomorphisms are dynamic in nature. They provided an algorithm to reduce transfer learning error by determining the order and regressors of these transfer learning maps.In \citet{gupta2017learning} trained invariant feature spaces using shared skills of morphologically different agents (simulated manipulators with various DOFs). These spaces were then used to share learned skill structure from one agent to others which accelerates the learning process of this skill in the latter. However, instead of facilitating the learning process across agents, our work aims to achieve zero-shot(no learning required) generalization of the learned skills. Working on a similar domain, recent work by \citet{hu2019skill} transferred acquired skills amongst morphologically different agents (MDAs). Compared to the work by \citet{gupta2017learning}, their experimental agents included considerably different morphologies (bipeds,single-legged hoppers, crawlers,etc.). They proposed a novel paired variational encoder-decoder to model morphology-invariant and morphology-dependent factors to expedite transfer learning. \citet{devin2017learning} decomposed the policies into task and robot-specific modules. These modules were trained in a mix and match fashion for both visual and non-visual tasks. Their method achieved zero-shot generalization to novel task and robot modules combinations which were not used during the training phase. This method however required the robot and task modules to be predetermined before training the entire ensemble of modules, which prevents this method from scaling to new robot modules without significant re-training.

	Among the plethora of works, CASNET is arguably most closely related to HCP-I introduced by \citet{chen2018hardware}. They proposed Hardware conditioned policies that use hardware information to generalize the policy network over robots with different kinematics and dynamics. Robot states were augmented with either explicit or implicit encodings of robot hardware and used as input to the policy network. However, they used zero-padding to generate fixed-length state-vectors for robots with different DOFs, which restricts the learned policy’s applicability to a predetermined limit of DOFs that cannot be extended without retraining the policy from scratch. In contrast, CASNET uses RNN based encoders and decoders to generate fixed-sized representation vectors from arbitrary sized state vectors and arbitrary sized action/output vectors from fixed-sized policy output.

	With CASNET, our aim is to take a meaningful step towards universal controllers which can optimally operate a plethora of different systems. Such steps have been taken for UAVs \citep{bulka2018universal} and gait generation of crawling robots \citep{malik2019generic}. However, with CASNET, we aim at providing a neural architecture for generating such controllers.

\bibliography{iclr2021_conference}
\bibliographystyle{iclr2021_conference}

\pagebreak
\begin{appendices}

	\section{Environment details}
	
	\subsection{Planer reachers}

	\begin{table}[H]
	\centering
	\caption{Reacher environments}
	\label{ReacherEnvironmentDetails}
	\begin{tabular}{|c|c|c|c|}
	\hline
	\multicolumn{4}{|c|}{Environment Details} \\
	\hline \hline
	Sr. No. & Name & DOFs & Link lengths \\ 
	\hline
	1 & Reacher\_10* & 1 & [10]\\
	\hline
	2 & Reacher\_11 & 1 & [15]\\
	\hline
	3 & Reacher\_12 & 1 & [09]\\
	\hline
	4 & Reacher\_20* & 2 & [12, 12]\\
	\hline
	5 & Reacher\_21 & 2 & [09, 14]\\
	\hline
	6 & Reacher\_22 & 2 & [13, 15]\\
	\hline
	7 & Reacher\_30* & 3 & [15, 17, 09]\\
	\hline
	8 & Reacher\_31 & 3 & [08, 11, 12]\\ 
	\hline
	9 & Reacher\_32 & 3 & [10, 10, 15]\\
	\hline
	10 & Reacher\_40* & 4 & [10, 16, 13, 09]\\
	\hline
	11 & Reacher\_41 & 4 & [13, 14, 07, 07]\\
	\hline
	12 & Reacher\_42 & 4 & [08, 15, 09, 11]\\
	\hline
	13 & Reacher\_50* & 5 & [10, 10, 10, 10, 10]\\
	\hline
	14 & Reacher\_51 & 5 & [15, 08, 09, 11, 13]\\
	\hline
	15 & Reacher\_52 & 5 & [10, 09, 12, 10, 14]\\
	\hline
	16 & Reacher\_60* & 6 & [10, 08, 15, 15, 10, 09]\\
	\hline
	17 & Reacher\_61 & 6 & [08, 09, 07, 13, 14, 07]\\
	\hline
	18 & Reacher\_62 & 6 & [10, 12, 08, 13, 07, 14]\\
	\hline
	\end{tabular}
	\end{table}

	Details of the 18 planer reacher environments used in experiments are shown in table \ref{ReacherEnvironmentDetails}. The environments used for training are marked with *. There are 3 environments with 1 to 6-DOFs with varying link lengths. Every actuator has the same torque limits and control range of -1 to +1 inclusive, which avoids the need to normalize actuator torque values for each environment. The state-space for each environment comprise of joint position, joint angular velocity and link-length for each actuator-link pair. The action space consist of the control values for the actuators. The objective for each environment is to reach a goal position randomly located with the robot's reach. The reward signal at every time-step is the negative sum of the distance between the robot's finger and the goal and squared average torque per actuator. Both these reward components are independent of the number of actuator-link pairs in reacher design. Each episode starts with reacher in fixed starting position (joint position corresponding to 0). An episode consist of 300 time-steps.

	\subsection{Crawling robots}
	Details about the crawling robot environments used in experiments are given in table \ref{CrawlerEnvironmentDetails} and table \ref{CrawlerEnvironmentDetails2}. The control range for each actuator is -1 to +1 inclusive. The state-space for each actuator-link pair consist of link-length, joint-range, actuator's axis or rotation and joint-position. Coxa's base location with respect to the center of mass (COM) of the robot is provided for each leg as encoded-specific input to the second encoder in the representation learner. Action-space consist of actuator control values. The objective for each environment is to walk in the forward direction (+y axis).The leg numbers are sequenced in clockwise fashion with respect to the z-axis (perpendicular to the surface pointing upwards). Each episode lasts 512 time-steps. At each time-step, reward is sum of movement reward, survival reward and torque-penalty. Movement rewards is propotional to COM's speed in forward direction and survival reward is +1 for each time-step the COM is not too close or far from the surface and 0 otherwise. Torque penalty is proportional to negative root mean square value of actuator torques per number of actuators.

	\begin{table}
	\centering
	\caption{Crawling robot environments. Note that L, C, F and T stand for Leg, Coxa, Femur and Tibia respectively. For example, C1 refers to the Coxa of Leg1.}
	\label{CrawlerEnvironmentDetails}
	\begin{tabular}{|c|c|c|c|c|c|}

	\hline
	\multicolumn{6}{|c|}{Environment Details} \\
	\hline\hline
	Sr. No. & Type & Suffix & Leg symmetry & DOFs per leg & Design modifications \\
	\hline
	
	1 & Quadruped & 10*	& Radial & 2 & None \\
	\hline
	2 & Quadruped & 11*	& Radial & 2 & 3-DOF in L1 \\
	\hline
	3 & Quadruped & 12	& Radial & 2 & 3-DOF in L1, L2 \\
	\hline
	4 & Quadruped & 13	& Radial & 3 & 2-DOF in L4 \\
	\hline
	5 & Quadruped & 14	& Radial & 3 & None \\
	\hline

	6 & Quadruped & 20	& Bilateral & 2 & None \\
	\hline
	7 & Quadruped & 21*	& Bilateral & 2 & 3-DOF in L1 \\
	\hline
	8 & Quadruped & 22*	& Bilateral & 2 & 3-DOF in L2 \\
	\hline
	9 & Quadruped & 23	& Bilateral & 2 & 3-DOF in L3 \\
	\hline
	10 & Quadruped & 24	& Bilateral & 2 & 3-DOF in L4 \\
	\hline
	11 & Quadruped & 25	& Bilateral & 3 & 2-DOF in L2, L4 \\
	\hline
	12 & Quadruped & 26*& Bilateral & 3 & None \\
	\hline

	13 & Quadruped & 31	& Bilateral & 2 & Modified C1,T1,T4 ranges \\
	\hline
	14 & Quadruped & 32*& Bilateral & 2 & Modified C3,C4,T2 ranges \\
	\hline
	15 & Quadruped & 33*& Bilateral & 2 & Modified C1,C2,C3 ranges \\
	\hline
	16 & Quadruped & 34	& Bilateral & 2 & Modified C3,C4,T4 ranges \\
	\hline
	17 & Quadruped & 35	& Bilateral & 2 & Modified C2,T1,T2 ranges \\
	\hline

	18 & Quadruped & 41	& Radial & 2 & Modified C1,T3,T4 lengths \\
	\hline
	19 & Quadruped & 42& Radial & 2 & Modified C2,C3,T3 lengths \\
	\hline
	20 & Quadruped & 43*& Radial & 2 & Modified C1,C2,T4 lengths \\
	\hline
	21 & Quadruped & 44*& Radial & 2 & Modified C1,T2,T4 lengths \\
	\hline
	22 & Quadruped & 45	& Radial & 2 & Modified C2,C3,T1 lengths \\
	\hline

	23 & Hexapod & 10*	& Radial & 2 & None \\
	\hline
	24 & Hexapod & 11*	& Radial & 2 & 3-DOF in L2 \\
	\hline
	25 & Hexapod & 12	& Radial & 2 & 3-DOF in L2, L6 \\
	\hline
	26 & Hexapod & 13	& Radial & 2 & 3-DOF in L2, L4, L6 \\
	\hline
	27 & Hexapod & 14	& Radial & 3 & 2-DOF in L3, L5 \\
	\hline
	28 & Hexapod & 15	& Radial & 3 & 2-DOF in L5 \\
	\hline
	29 & Hexapod & 16*	& Radial & 3 & None \\
	\hline

	30 & Hexapod & 20	& Bilateral & 2 & None \\
	\hline
	31 & Hexapod & 21*	& Bilateral & 2 & 3-DOF in L1 \\
	\hline
	32 & Hexapod & 22*	& Bilateral & 2 & 3-DOF in L1, L3 \\
	\hline
	33 & Hexapod & 23	& Bilateral & 3 & 2-DOF in L2, L4, L6 \\
	\hline
	34 & Hexapod & 24	& Bilateral & 3 & 2-DOF in L4, L6 \\
	\hline
	35 & Hexapod & 25*	& Bilateral & 3 & 2-DOF in L6 \\
	\hline
	36 & Hexapod & 26	& Bilateral & 3 & None \\
	\hline
	
	37 & Hexapod & 31	& Radial & 2 & Modified C1,C6,T2 ranges \\
	\hline
	38 & Hexapod & 32*	& Radial & 2 & Modified C3,T3,T4 ranges \\
	\hline
	39 & Hexapod & 33*	& Radial & 2 & Modified T1,T3,T5 ranges \\
	\hline
	40 & Hexapod & 34	& Radial & 2 & Modified C2,C5,T6 ranges \\
	\hline
	41 & Hexapod & 35	& Radial & 2 & Modified C2,T3,T4 ranges \\
	\hline

	42 & Hexapod & 41	& Bilateral & 2 & Modified C1,C2,C6 lengths \\
	\hline
	43 & Hexapod & 42	& Bilateral & 2 & Modified C6,T3,T5 lengths \\
	\hline
	44 & Hexapod & 43*	& Bilateral & 2 & Modified C3,C4,T4 lengths \\
	\hline
	45 & Hexapod & 44*	& Bilateral & 2 & Modified C5,T1,T6 lengths \\
	\hline
	46 & Hexapod & 45	& Bilateral & 2 & Modified C1,T2,T5 lengths \\
	\hline

	\end{tabular}
	\end{table}

	\begin{table}
	\centering
	\caption{Crawling robot environments with multiple DOVs.}
	\label{CrawlerEnvironmentDetails2}
	\begin{tabular}{|c|c|c|c|c|l|}

	\hline
	\multicolumn{6}{|c|}{Environment Details} \\
	\hline\hline
	Sr. No. & Type & Suffix & Leg symmetry & DOFs per leg & Design modifications \\
	\hline

	1 & Quadruped & 	51	& Radial & 2 & \makecell[l]{- L1 base location\\- T1 length\\- C3, T4 ranges} \\
	\hline
	2 & Quadruped & 	52	& Bilateral & 2 & \makecell[l]{- 3 DOF L3\\- F3 length\\- C2, T1 ranges} \\
	\hline
	3 & Quadruped & 	53	& Radial & 3 & \makecell[l]{- 2 DOF L4\\- F3 length\\- L3, L4 base location\\- F1 range} \\
	\hline
	4 & Quadruped & 	54	& Radial & 3 & \makecell[l]{- L3 base location\\- F1 length\\- T2 range} \\
	\hline
	5 & Quadruped & 	55	& Radial & 3 & \makecell[l]{- 2 DOF L1,L3\\- F4 length\\- C2, T1, F4 ranges} \\
	\hline
	6 & Hexapod & 		51	& Radial & 2 & \makecell[l]{- 3 DOF L2, L6\\- T2, C4 lengths\\- C1, F6 ranges} \\
	\hline
	7 & Hexapod & 		52	& Radial & 2 & \makecell[l]{- 3 DOF L2\\- F2 length\\- T5 range} \\
	\hline
	8 & Hexapod &		53	& Radial & 2 & \makecell[l]{- L5 base location\\- F2 length\\- C1, T1 range} \\
	\hline
	9 & Hexapod & 		54	& Bilateral & 3 & \makecell[l]{- L4 base location\\- 2 DOF L2,L4,L6\\- F3 length\\- C3, T3 range} \\
	\hline
	10 & Hexapod & 		55	& Bilateral & 3 & \makecell[l]{- L6 base location\\- C3, T3 range\\- F2 range} \\
	\hline

	\end{tabular}
	\end{table}

	\section{Hyperparameters}

	\begin{table}[H]
	\centering
	\caption{Planer reachers}
	\label{ReacherHyperparameters}
	\begin{tabular}{|c|c|}
	\hline
	Learning algorithm	&	Proximal Policy Optimization	\\
	\hline
	Learning rate 		&	0.0003							\\
	\hline
	Total episodes		&	2000							\\
	\hline
	Batch size			&	4								\\
	\hline
	Clip-range			&	0.2								\\
	\hline
	Gamma				&	0.99							\\
	\hline
	Lambda				&	0.95							\\
	\hline
	\end{tabular}
	\end{table}

	\begin{table}[H]
	\centering
	\caption{Crawling robots}
	\label{CrawlerHyperparameters}
	\begin{tabular}{|c|c|}
	\hline
	Learning algorithm	&	Soft Actor Critic	\\
	\hline
	Learning rate 		&	0.0003				\\
	\hline
	Total num-steps		&	200000				\\
	\hline
	Batch size			&	256					\\
	\hline
	Polyak 				&	0.05				\\
	\hline
	Gamma				&	0.99				\\
	\hline
	Alpha				&	0.2					\\
	\hline
	\end{tabular}
	\end{table}

	\begin{table}[H]
	\centering
	\caption{Manipulators}
	\label{ManipulatorHyperparameters}
	\begin{tabular}{|c|c|}
	\hline
	Learning algorithm		&	Deep Deterministic Policy Gradient	\\
	\hline
	Learning rate 			&	0.0001								\\
	\hline
	Total num-steps			&	625000								\\
	\hline
	Update per episode		&	40									\\
	\hline
	Num-steps per episode	&	50									\\
	\hline
	Batch size				&	64									\\
	\hline
	Polyak 					&	0.95								\\
	\hline
	Gamma					&	0.98								\\
	\hline
	Replay\_K				&	4									\\
	\hline
	\end{tabular}
	\end{table}

\end{appendices}

\end{document}